\documentclass{article}

\usepackage{microtype}
\usepackage{graphicx}
\usepackage{subcaption}
\usepackage{booktabs} 
\usepackage{multirow} 

\usepackage{hyperref}

\usepackage[accepted]{icml2026}

\usepackage{amsmath}
\usepackage{amssymb}
\usepackage{mathtools}
\usepackage{amsthm}

\usepackage[capitalize,noabbrev]{cleveref}

\theoremstyle{plain}

\theoremstyle{definition}

\theoremstyle{remark}

\usepackage[disable,textsize=tiny]{todonotes}

\icmltitlerunning{Two-Stage Fine-Tuning for Protein Sequence Generation with Targeted Amino-Acid Composition}

\begin{document}

\twocolumn[
  \icmltitle{Two-Stage Fine-Tuning for Protein Sequence Generation with Targeted Amino-Acid Composition}



  \icmlsetsymbol{equal}{*}

  \begin{icmlauthorlist}
    \icmlauthor{Violeta Basten-Romero}{bsc,sch}
    \icmlauthor{Rubén Muñoz-Tafalla}{bsc}
    \icmlauthor{Anna María Díaz-Rovira}{bsc}
    \icmlauthor{Bertran Miquel-Oliver}{bsc,sch}
    \icmlauthor{Isaac Filella-Merce}{bsc,NB}
    \icmlauthor{Víctor Guallar}{bsc,NB,ICREA}
  \end{icmlauthorlist}

  \icmlaffiliation{bsc}{Barcelona Supercomputing Center (BSC), Plaça d’Eusebi Güell, 1-3, 08034, Barcelona, Spain}
  \icmlaffiliation{sch}{Department of Bioinformatics, School of Mathematics and Statistics (FME), Universitat Politècnica de Catalunya - BarcelonaTech (UPC), 08034 Barcelona, Spain}
  \icmlaffiliation{NB}{Nostrum Biodiscovery S.L., Av. de Josep Tarradellas, 8-10, 3-2, 08029, Barcelona, Spain}
  \icmlaffiliation{ICREA}{ICREA, Pg. Lluis Companys 23, 08010, Barcelona, Spain}

  \icmlcorrespondingauthor{Victor Guallar}{victor.guallar@bsc.es}

  \icmlkeywords{protein design, language models, reinforcement learning, composition control, reward-weighted regression}

  \vskip 0.3in
]



\printAffiliationsAndNotice{}  

\begin{abstract}
Protein language models are standard priors for biological sequence generation, but steering them toward explicit distributional design targets remains largely unexplored. We study a constrained protein generation problem in which sequences must match a desired amino-acid (AA) composition profile while preserving plausible sequence statistics and diversity. The motivating application is synthetic feed protein design, where the AA composition of dietary proteins directly determines their nutritional value. We propose a two-stage pipeline in which domain-adaptive fine-tuning (FT) on an in-domain protein dataset is followed by iterative reward-weighted FT via reinforcement learning (RL) anchored against the FT model as a frozen reference. We evaluate the pipeline on two AA compositions and find that FT brings the average composition close to the target, while the subsequent RL enforces specific sequence constraints that FT alone cannot satisfy. We additionally evaluate the design choices of the proposed composition reward term against two baselines and an ablated variant, isolate the contribution of each training stage, and verify that AA composition alignment is achieved without degrading sequence quality.
\end{abstract}

\section{Introduction}

Protein language models (PLMs) such as ProtGPT2~\citep{ferruz2022protgpt2}, ProGen2~\citep{nijkamp2023progen2}, and RITA~\citep{hesslow2022rita} have become standard priors for \textit{de novo} sequence generation. In most design settings, however, plausibility is not enough: sequences must satisfy an explicit external objective. Accordingly, PLMs can be steered toward a range of design objectives, including specific family fold~\citep{madani2023progen}, predicted structural confidence~\citep{stocco2024dpoplm,subramanian2024rlpf}, enzymatic activity~\citep{munsamy2024zymctrl,stocco2024dpoplm}, thermostability and binding fitness~\citep{widatalla2024proteindpo}, and antimicrobial activity~\citep{cao2025apexamphion}. These objectives are pursued through steering strategies that range from prompt-based conditioning, which uses conditioning tags the model was trained to recognize~\citep{madani2023progen,munsamy2024zymctrl}, to supervised fine-tuning (FT) ~\citep{madani2023progen} and, more recently, reward-guided FT via reinforcement learning (RL)~\citep{cao2025apexamphion,stocco2024dpoplm,subramanian2024rlpf}.

These steering strategies frame the design objective as either a categorization (i.e., predicting membership to a predefined class) or a regression (i.e., optimizing a scalar derived from experimental measurements or \textit{in silico} scoring). They are thus not directly suited to design objectives that require matching a distributional target profile, such as a charge profile, a hydrophobicity pattern, or an amino-acid (AA) composition. Generating proteins with controlled AA compositions could enable applications across biotherapeutics and immunology, biomaterials, and nutrition. Focusing on nutritional applicability, the design objective of this work is synthetic-feed protein design, in which candidate proteins should not only match a prescribed AA composition that reflects an idealized nutritional profile but also remain synthesizable and diverse \citep{cambralopez2022,ravindran2013}. The nutritional value of a protein source is largely determined by its AA composition. In practice, dietary proteins often have AA compositions that do not fully align with the organism's nutritional requirements, reducing protein digestibility and nitrogen retention \citep{emmert1997ideal}. An idealized nutritional profile thus refers to a target AA composition that maximizes coverage of these dietary requirements.

To this end, we propose a two-stage pipeline for post-training alignment under an explicit AA composition objective. Starting from a pretrained PLM (ProtGPT2), the first stage applies domain-adaptive FT to a subset of natural proteins whose AA composition is closest to the target composition, thereby anchoring the prior near the desired region of sequence space. The second stage applies iterative reward-weighted FT via RL, in which at each iteration we generate candidates, score them with a composition reward formulation, filter for length and diversity, and update the policy. We evaluate the pipeline on two distinct target compositions, including a published reference \citep{cambralopez2022} and an in-domain idealized composition. Finally, we report the contribution of each training stage, an evaluation of the design choices of our composition reward formulation, and an analysis of preserved sequence quality.

\section{Related Work}

Our work sits within a broader literature on controlled generation for protein sequences. Recent work has increasingly used reward-guided FT via RL to steer pretrained PLMs toward design objectives that go beyond sequence plausibility. \citet{stocco2024dpoplm} introduces DPO\_pLM, applying direct preference optimization~\citep{rafailov2023dpo} to autoregressive PLMs against oracles such as ESMFold pLDDT and the CLEAN enzyme classifier. \citet{widatalla2024proteindpo} applies DPO to ESM-IF1 using experimental thermostability measurements, converting scalar stability labels such as $\Delta G$ or $\Delta\Delta G$ into paired, ranked, or weighted preference objectives. \citet{cao2025apexamphion} fine-tunes ProGen2-XL with proximal policy optimization against a composite reward combining a learned minimum-inhibitory-concentration classifier and physicochemical descriptors, designing antimicrobial peptides validated experimentally. \citet{subramanian2024rlpf} uses RL on PLMs with structural-confidence rewards distilled from ESMFold. These methods align PLMs against external \emph{scalar, ordinal, or categorical} oracles, such as structure-confidence scores, enzyme-class predictions, thermostability measurements, or antimicrobial-activity labels. Our setting is complementary: the reward directly measures alignment to a target AA composition and is computable analytically from the sequence itself, requiring no external model or experimental measurement.

Our objective is closest to reward-weighted and advantage-weighted regression~\citep{peters2007rwr,peng2019awr}, with the frozen reference model as a  KL trust region in the spirit of DPO~\citep{rafailov2023dpo} and PPO. Rather than constructing preference pairs as in DPO, we use a softmax reward-weighting on the per-batch candidate pool, which naturally handles continuous-valued rewards without explicit pair construction. The loss is given in \S\ref{sec:method:rl}.

\section{Methods}

\subsection{Problem Formulation}

Let $p(s)\in\mathcal{S}_{20}$ denote the empirical AA frequency vector of a protein sequence $s$, where $\mathcal{S}_{20}=\{x\in\mathbb{R}_{\geq 1}^{20}:\sum_i x_i=1\}$ is the probability simplex over the twenty canonical AA, and let $q\in\mathcal{S}_{20}$ be a target composition. We want to adapt a pretrained PLM so that it generates sequences satisfying $p(s)\approx q$  while preserving sequence plausibility, valid length, and pairwise diversity.

\subsection{First Stage: Base Model and Domain-Adaptive FT}
\label{sec:method:ft}

Pretraining on in-domain data is a standard recipe for domain adaptation~\citep{gururangan2020dont}. We use it as a composition-conditioned adaptation, anchoring ($\pi_\text{ref}$) near a region of sequence space that is both biologically plausible and compositionally similar to $q$. Starting from the UniProtKB/TrEMBL release~\citep{uniprot2023} (the unreviewed, automatically annotated portion of UniProt; downloaded in FASTA format from the UniProt Consortium FTP repository, $\sim$$2.5{\times}10^{8}$ sequences), we apply three filters: (i) a length filter $100$-$500$ AA (retains $\sim$$1.8{\times}10^{8}$ sequences); (ii) a cosine-similarity filter against $q$ at a threshold $\geq 0.95$, which retains $\sim$$2.5{\times}10^{5}$ sequences whose own composition resembles the target composition; and (iii) a sequence-identity filter at $<70\%$ pairwise identity to remove near-duplicates and avoid bias toward over-represented families, leaving $\sim$$1.0{\times}10^{5}$ sequences. The same pipeline is applied to two target compositions ($q_\text{A}$ and $q_\text{B}$; \S\ref{sec:exp:targets}), yielding two distinct FT datasets. The resulting FT dataset contains natural proteins that are both biologically plausible and compositionally close to $q$, providing a training signal that biases the base model toward the target composition before any reward-guided RL optimization.

The base model is ProtGPT2~\citep{ferruz2022protgpt2}. We perform causal-LM domain-adaptive FT on the FT dataset and take the resulting checkpoint (FT prior) as the frozen reference policy $\pi_\text{ref}$ for all subsequent RL runs (hyperparameters in App.~\ref{app:repro}). 

\subsection{Second Stage: Reward-Weighted RL}
\label{sec:method:rl}

Given the frozen reference policy $\pi_\text{ref}$, we optimize a trainable policy $\pi_\theta$ with a reward-weighted log-ratio objective. For a batch of sampled sequences $\{s_i\}$, we compute scalar rewards $r(s_i)$, standardize and clip them within the batch to obtain $\tilde{r}_i$, and convert them to weights $w_i =\mathrm{softmax}(\tilde r_i)$, so that sequences with higher rewards contribute more to the update. The objective is
\begin{equation}
\begin{split}
\mathcal{L}(\theta) =
\;&-\sum_i w_i\,\eta\bigl(\log\pi_\theta(s_i) - \log\pi_\text{ref}(s_i)\bigr)
\; \\
&+\; \lambda_\text{KL}\,\mathrm{KL}\!\bigl(\pi_\theta\,\|\,\pi_\text{ref}\bigr).
\end{split}
\label{eq:loss}
\end{equation}
where $\eta$ is a log-ratio scale factor. This is related to reward-weighted and advantage-weighted regression~\citep{peters2007rwr,peng2019awr}, with $\pi_\text{ref}$  playing a role analogous to that in DPO~\citep{rafailov2023dpo}. The Kullback-Leibler term $\mathrm{KL}(\pi_\theta\|\pi_\text{ref})$ measures how far $\pi_\theta$ has drifted from $\pi_\text{ref}$. Penalizing it acts as a trust region that prevents $\pi_\theta$ from exploiting reward shortcuts at the expense of sequence plausibility.

Each iteration $t$: (i) generate a candidate pool of sequences with stochastic decoding; (ii) generate a parallel \emph{diversity-pulse} pool with hotter decoding (higher temperature and $\mathrm{top\text{-}p}$; \S\ref{app:divboost}); (iii) score both with $r(s)$; (iv) filter sequences by length, de-duplication, and pairwise identity below $0.85$ using MMseqs2~\citep{steinegger2017mmseqs2}; (v) re-inject a fraction of the diversity-pulse pool into the candidate pool, with the re-injection fraction increasing across training to counteract late-iteration mode collapse; (vi) construct reward-weighted batches and update $\pi_\theta$ under Eq.~\ref{eq:loss}.

The reward combines a composition term and a length term,
\begin{equation}
r(s) = w_c\,\mathrm{Comp}(s,q) + w_l\,\mathrm{Len}(s),
\end{equation}
with static weights $(w_c, w_l) = (0.97, 0.03)$ across all runs. The composition term is the primary signal, while the length term prevents collapse to extremes, motivating the asymmetric weighting. These weights were chosen as a working default after a small exploration. No controlled sweep was performed, so we make no claim of optimality. $\mathrm{Len}(s)$ is a piecewise-linear shaping term that is non-zero on $[70, 400]$ AA with a peak plateau on $[110, 250]$ (App.~\ref{app:rewards}). This range used at training does not have to coincide with the FT dataset length filter $[100,500]$ AA. The length term and weights are identical across all reward variants analyzed in this work, and only $\mathrm{Comp}(s,q)$ differs.

We propose a \emph{differentiated} composition term as our primary $\mathrm{Comp}(s,q)$ term and compare it against one ablation and two baselines at matched compute to assess the contribution of its design choices. The \emph{differentiated} composition term uses an asymmetric per-residue kernel (penalizing deficits in essential AA, those that cannot be synthesized by the organism and must be obtained through diet, more harshly than excesses), two residue pools whose members are treated as biochemically interchangeable (a sulfur pool (Met/Cys) and an aromatic-precursor pool (Phe/Tyr)), and a zero-target residue amplifier (a term that up-weights residues whose target frequency is zero). As an ablation, we evaluate a \emph{symmetric} variant of the \emph{differentiated} term in which the asymmetric per-residue weighting is replaced by uniform absolute deviations. As baselines, we additionally evaluate a \emph{cosine} similarity term and a \emph{global-deviation} $L_1$ composition term, both of which lack the per-residue structure of the \emph{differentiated} composition term. Full formulas and hyperparameters App.~\ref{app:rewards}.

Each composition term has a sharpness coefficient $\beta$ inside an outer $\exp(-\beta\,[\cdot])$ that maps the per-sequence composition error to a bounded reward, with $\beta$ ramping during training to progressively sharpen the reward signal. A fixed reference value $\beta_\text{ref}$ is used at evaluation, applied to the \emph{differentiated} term so that all variants are compared on a common scoring function (full per-variant ramps in App.~\ref{app:rewards}; loop hyperparameters in App.~\ref{app:repro}).

\section{Experimental setup}
\label{sec:exp}

\subsection{Target AA Compositions}
\label{sec:exp:targets}

$q_\text{A}$ (primary): an experimentally refined poultry-feed AA composition provided by a project partner (details withheld for proprietary reasons). It includes two low-frequency residues (one at exactly zero). We anonymized the residue labels as $\mathrm{aa}_1,\ldots,\mathrm{aa}_{20}$ throughout (sorted by descending target frequency).

$q_\text{B}$ (published reference): a poultry-feed AA composition derived from~\citet{cambralopez2022}, with twenty non-zero target frequencies. Used as a generalization probe.

\begin{table*}[hbt!]
\caption{Pipeline comparison on the primary target $q_\text{A}$ and the published reference target $q_\text{B}$. All values are pool means at the run's best iteration (selected by mean composition score at $\beta_\text{ref}{=}20$). The ``RL ($n{=}\cdot$)'' rows aggregate all seeds of the \emph{differentiated} composition term on the corresponding target (mean$\,\pm\,$std across seeds for JSD and Comp.), and the ``RL (best run)'' rows report the single seed with the lowest pool-mean JSD on that target.}
\label{tab:pipeline}
  \begin{center}
    \begin{small}
      \begin{sc}
        \begin{tabular}{lcccc}
            \toprule
            Stage & Target & Comp.$\uparrow$ & JSD$\downarrow$ & $N_{\pm 30}\uparrow$ \\
            \midrule
            Base ProtGPT2                 & $q_\text{A}$ & 0.007 & 0.247 & \phantom{0}3.70 \\
            + Domain-adapt. FT                   & $q_\text{A}$ & 0.126 & 0.059 & \phantom{0}9.17 \\
            + RL ($n{=}30$)               & $q_\text{A}$ & $0.397{\pm}0.210$ & $0.032{\pm}0.020$ & 12.34 \\
            + RL (best run)               & $q_\text{A}$ & \textbf{0.822} & \textbf{0.0044} & \textbf{17.40} \\
            \midrule
            Base ProtGPT2                 & $q_\text{B}$ & 0.054 & 0.202 & \phantom{0}3.71 \\
            + Domain-adapt. FT                   & $q_\text{B}$ & 0.207 & 0.051 & \phantom{0}8.25 \\
            + RL ($n{=}10$)               & $q_\text{B}$ & $0.566{\pm}0.123$ & $0.021{\pm}0.008$ & 12.21 \\
            + RL (best run)               & $q_\text{B}$ & \textbf{0.830} & \textbf{0.0008} & \textbf{20.00} \\
            \bottomrule
        \end{tabular}
      \end{sc}
    \end{small}
  \end{center}
  \vskip -0.1in
\end{table*}

\subsection{Runs} 
\label{sec:exp:runs}

We run all four composition term variants of \S\ref{sec:method:rl} on each target at matched compute and shared seeds. On $q_\text{A}$, each variant is run on $30$ seeds; on $q_\text{B}$, each variant is run on $10$ seeds. All other recipe knobs (number of iterations, optimizer, KL schedule, $\beta$ ramp, candidates per iteration, and the length and identity filters) are identical across variants and across targets. This gives $4{\times}30 + 4{\times}10 = 160$ RL runs in total, plus base ProtGPT2 (no further training) and a domain-adaptive FT checkpoint per target composition, evaluated as four baseline rows in Table~\ref{tab:pipeline}.

\subsection{Metrics}

We evaluate each run at the iteration whose candidate pool has the highest mean composition score, recomputed at fixed $\beta_\text{ref}{=}20$ through the \emph{differentiated} composition term (App.~\ref{app:rewards}), so that selection is independent of each run's $\beta$ schedule.

We report the \emph{Jensen-Shannon divergence} $\mathrm{JSD}(p,q)$ between the empirical and target composition, the \emph{tolerance count} $N_{\pm 30}(p,q)$, defined as the number of residues whose empirical frequency falls within a $30\%$ relative window of their target frequency, and the composition score at $\beta_\text{ref}$. All three are computed per sequence and then averaged over the candidate pool of the selected iteration. We therefore report pool means for each run, and, for the multi-seed ''RL'' rows, mean$\,\pm\,$std across seeds of these per-run pool means. Pool sizes range from $\sim$$700$ to $\sim$$2000$ sequences depending on iteration, after length and identity filtering. We additionally report three indicators measuring the fraction of sequences in the candidate pool that satisfy a given sequence constraint: \emph{essential-residue coverage} (fraction of sequences in which the ten essential AA each reach at least half of their target frequency), \emph{pool tolerance} (both interchangeable pools within $\pm 30\%$ of their pool target), and \emph{low-target compliance} (at most two occurrences of any residue whose target frequency is at or near zero). The low-target compliance metric is reported only on $q_\text{A}$ because $q_\text{B}$ has no zero or near-zero frequency residues. 

Sequence quality (\S\ref{sec:results:quality}) is summarized by NetSolP-predicted solubility~\citep{thumuluri2022netsolp}, base ProtGPT2 log-PPL, ESM-2 pPPL, mean length, and intra-pool similarity. Per-variant means use $95\%$ percentile bootstrap CIs over the seed count.

\begin{table*}[hbt!]
\caption{Pipeline comparison of  sequence constraint indicators on the primary target $q_A$ and the published reference target $q_B$ fraction of sequences in the candidate pool satisfying each indicator). The ``RL ($n{=}\cdot$)'' rows aggregate all seeds of the corresponding target, and the ``RL (best run)'' rows report the single seed with the lowest pool-mean JSD on that target.}
\label{tab:constraints}
  \begin{center}
    \begin{small}
      \begin{sc}
        \begin{tabular}{lcccc}
            \toprule
            Stage & Target & Essent.$\uparrow$ & Pools$\uparrow$ & Low-tgt.$\uparrow$ \\
            \midrule
            Base ProtGPT2          & $q_\text{A}$ & 0.004 & 0.060 & 0.226 \\
            + Domain-adapt. FT            & $q_\text{A}$ & 0.212 & 0.276 & 0.098 \\
            + RL (mean, $n{=}30$)  & $q_\text{A}$ & 0.470 & 0.553 & 0.602 \\
            + RL (best run)        & $q_\text{A}$ & \textbf{1.000} & \textbf{1.000} & \textbf{1.000} \\
            \midrule
            Base ProtGPT2          & $q_\text{B}$ & 0.030 & 0.032 & -- \\
            + Domain-adapt. FT            & $q_\text{B}$ & 0.160 & 0.224 & -- \\
            + RL (mean, $n{=}10$)  & $q_\text{B}$ & 0.724 & 0.672 & -- \\
            + RL (best run)        & $q_\text{B}$ & \textbf{1.000} & \textbf{1.000} & -- \\
            \bottomrule
        \end{tabular}
      \end{sc}
    \end{small}
  \end{center}
  \vskip -0.1in
\end{table*}

\section{Results}
\label{sec:results}

\subsection{Domain Adaptation Moves the Base Model Toward the Target AA Composition}
\label{sec:results:ft}

We first examine the effect of domain-adaptive FT before any reward-weighted RL is applied. Independent FT runs on the respective target-specific FT datasets produce a large shift toward their target compositions (Table~\ref{tab:pipeline}).

On $q_\text{A}$, held-out perplexity on the FT eval split (a held-out subset of the composition-filtered UniProt sequences, \S\ref{sec:method:ft}, App.~\ref{app:repro}) drops by an order of magnitude ($5570.6 \to 507.3$). Concurrently, JSD drops from $0.247$ to $0.059$ bits, the tolerance count $N_{\pm 30}$ rises from $3.70$ to $9.17$, and the composition score rises from $0.007$ to $0.126$ (Table \ref{tab:pipeline}). Because this eval split is drawn from the composition-filtered UniProt subset that conditions the FT (\S\ref{sec:method:ft}), the held-out perplexity primarily reflects domain adaptation rather than generic protein-likeness. For an independent protein-likeness proxy, we use ESM-2 (150M) pseudo-perplexity (pPPL)~\citep{lin2023esm2}, which we report in full in \S\ref{sec:results:quality} (Table~\ref{tab:esm2}) and which shows that FT is slightly worse than base ProtGPT2. On $q_\text{B}$ the picture is qualitatively the same (Table~\ref{tab:pipeline}): JSD drops from $0.202$ to $0.051$, $N_{\pm 30}$ rises from $3.71$ to $8.25$, and the composition score from $0.054$ to $0.207$.

One qualification follows. The average composition is closer to $q$, but sequence constraint indicators remain weak after FT (essential-residue coverage $0.212$, pool tolerance $0.276$, low-target compliance $0.098$ on $q_\text{A}$, with analogous gaps on $q_\text{B}$; Table~\ref{tab:constraints}). This is expected, as the FT dataset is built from natural proteins selected by AA composition cosine similarity (\S\ref{sec:method:ft}), and the unconstrained likelihood objective cannot enforce sharp residue constraints that diverge from typical natural composition. The reward-weighted RL stage closes exactly this gap (\S\ref{sec:results:rl}). Whether the FT stage is necessary for this gap closure is examined as an ablation in \S\ref{sec:results:ablation:noft}.

\subsection{Reward-Weighted RL Closes Sequence Constraints Gap}
\label{sec:results:rl}

Having established that domain-adaptive FT improves average composition but leaves sequence constraints largely unsatisfied, we now show that adding the RL stage on top of FT closes the remaining gap. Using the \emph{differentiated} term on $q_\text{A}$ results in a mean JSD of $0.032 \pm 0.020$ across seeds ($n=30$), compared to $0.059$ for FT alone, that is, a roughly $2\times$ reduction (Table~\ref{tab:pipeline}). The seed distribution is right-skewed: the best run reaches $\mathrm{JSD}=0.0044$ a composition score of $0.822$, $N_{\pm 30}=17.4$ (Table~\ref{tab:pipeline}), and all sequence constraints indicators (low-target compliance, essential-residue coverage, pool tolerance) are saturated at $1.00$, while FT and base ProtGPT2 leave most sequences out of compliance (Table~\ref{tab:constraints}). The mean-across-seeds row is intermediate, with typical seeds satisfying roughly half of each sequence constraint indicator. Five of the $30$ seeds reach $\mathrm{JSD}<0.01$. The best-seed numbers are operationally relevant because the best runs are the ones whose candidates will be selected for synthesis. The full top-5 by JSD and by composition score is reported in App.~\ref{app:top5}. On $q_\text{B}$ the \emph{differentiated} term reaches a mean $\mathrm{JSD}=0.021\pm 0.008$ (n=10). The best run achieves $\mathrm{JSD} = 0.0008$, a composition score of 0.830, $N_{\pm 30}=20.0$ and all applicable sequence constraints indicators saturated at 1.00 (Table~\ref{tab:pipeline}). The $q_\text{B}$ best run is closer to its target than the $q_\text{A}$ best run, but $q_\text{B}$ is a strictly easier task: it has no zero-frequency residues, so the zero-target residue amplifier that dominates the \emph{differentiated} composition term on $q_\text{A}$ does not apply. The two ''best'' numbers are therefore not directly comparable. 

\begin{figure}[ht!]
  \vskip 0.2in
  \begin{center}
  \centerline{\includegraphics[width=\columnwidth]{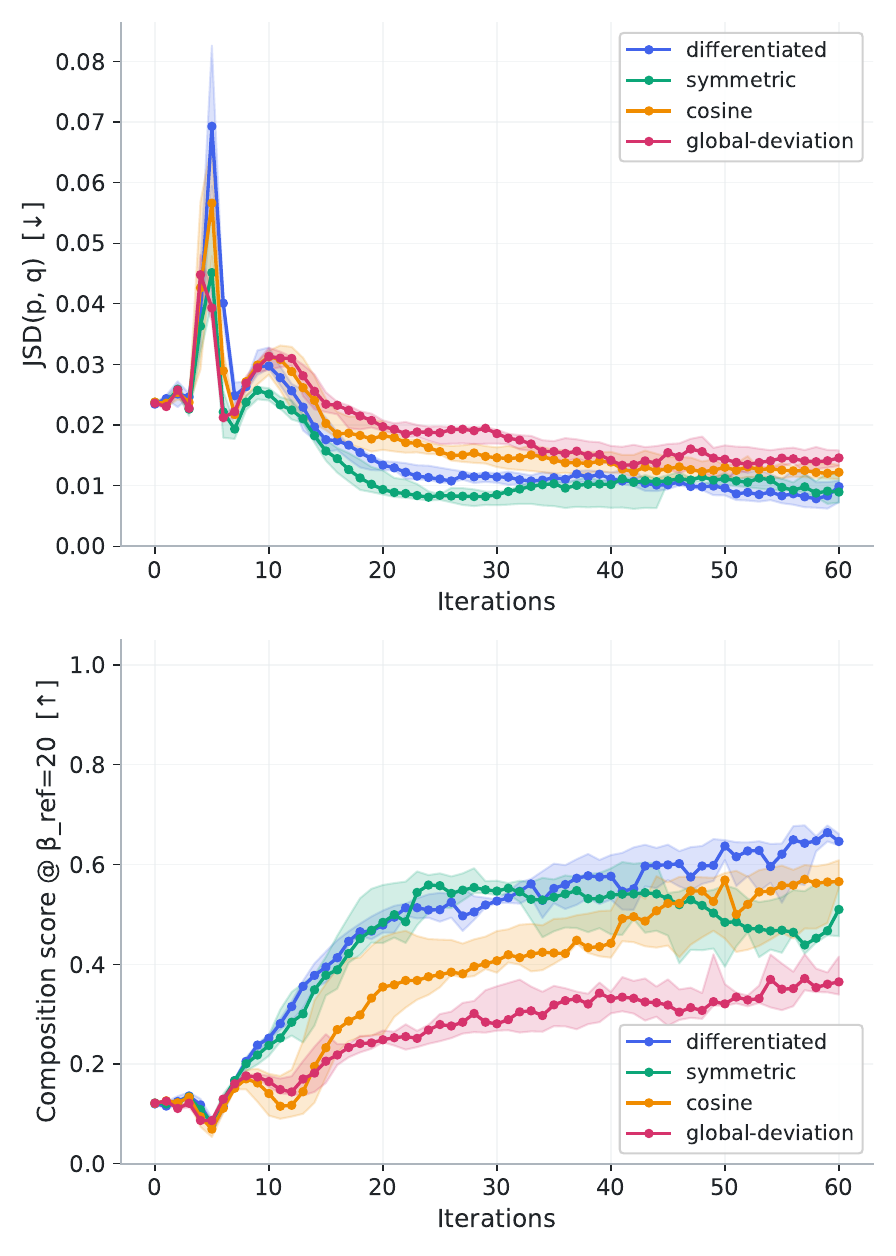}}
    \caption{Training dynamics on $q_\text{A}$. Each curve is the median across the top six seeds per composition term variant, with the IQR shaded. (Top) Mean JSD against the target across iterations. (Bottom) Composition score evaluated at a fixed reference temperature $\beta_\text{ref}{=}20$ using the \emph{differentiated} composition term for all four variants.}
\label{fig:dynamics}
\end{center}
\end{figure}

The fixed-$\beta_\text{ref}$ composition score rises monotonically across iterations (Fig.~\ref{fig:dynamics}, bottom) for the median of the top six seeds of every composition term variant. Whether the RL gains over FT can be matched by best-of-N selection from the FT prior alone is examined as an ablation in \S\ref{sec:results:ablation:bestofn}.

\subsection{Per-Residue Calibration}

\begin{figure}[htb!]
  \vskip 0.2in
  \begin{center}
  \centerline{\includegraphics[width=\columnwidth]{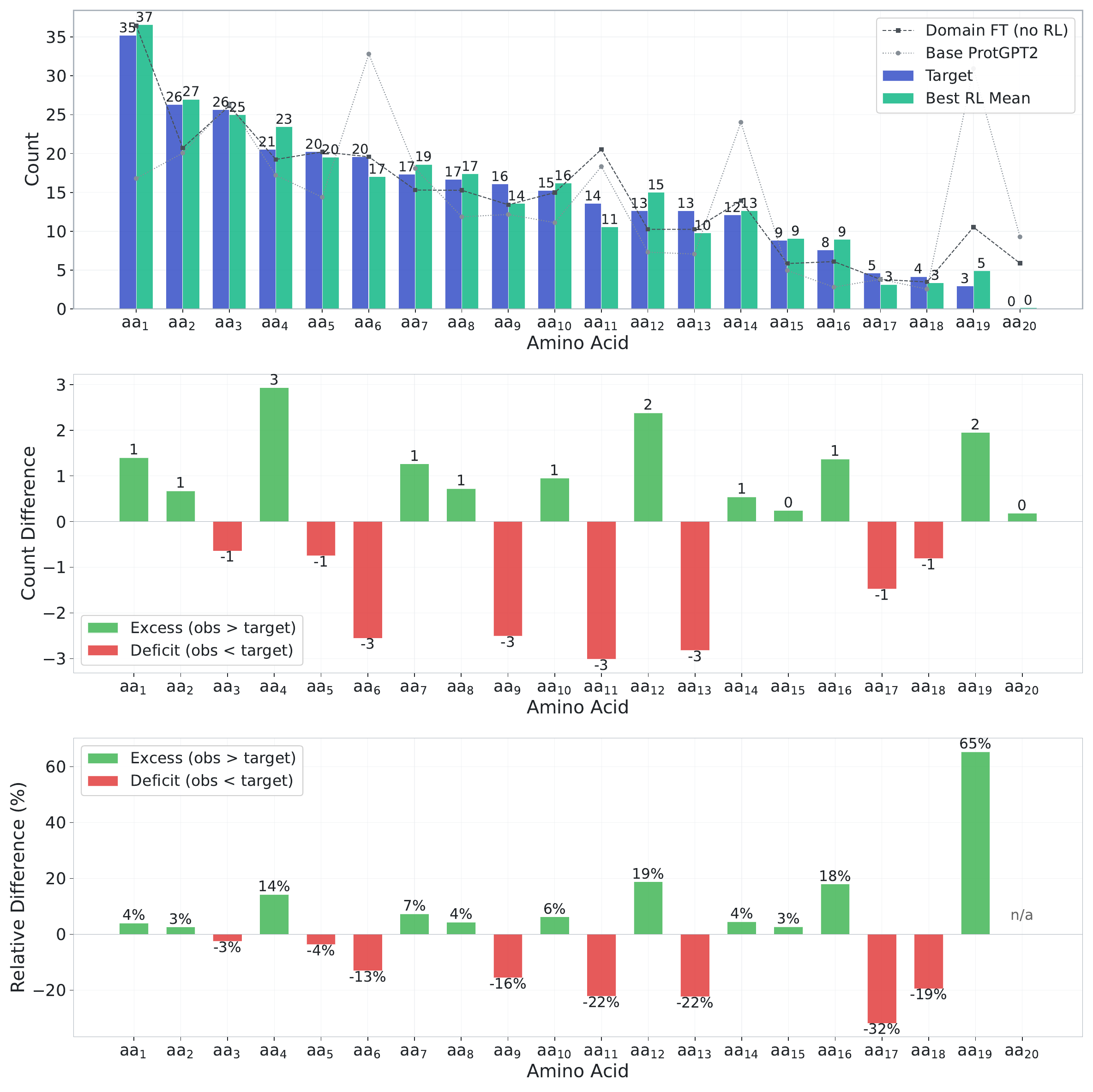}}
\caption{Per-residue calibration on $q_\text{A}$ (residues anonymized as $\mathrm{aa}_1,\ldots,\mathrm{aa}_{20}$, sorted by descending $q$). Counts are pool-mean frequencies $p_i$ rescaled to a common reference length $L{=}292$ AA (the rounded mean sequence length of the best-RL pool). (Top) Target counts (blue) vs.\ best-RL counts (turquoise); Domain-adaptive FT and base ProtGPT2 are overlaid as dashed and dotted lines. (Middle) Signed count residual (observed vs. target) for the best-RL run. (Bottom) Signed relative residual $(p-q)/q\times 100\%$ for the best-RL run (length-independent), the $q{=}0$ residue is marked n/a.}
\label{fig:calibration}
\end{center}
\end{figure}

Figure~\ref{fig:calibration} shows the per-residue composition calibration of the best RL run relative to the FT prior and base ProtGPT2, revealing where each training stage contributes most. The top panel shows observed vs. target counts for every residue (anonymized as $\mathrm{aa}_1,\ldots,\mathrm{aa}_{20}$, sorted by descending target frequency). The highest-frequency target residues are already well-matched after FT, with RL providing only modest additional sharpening. Intermediate-frequency residues (positions $\sim$5-15) are where RL contributes the most, often pulling absolute frequency residuals $|p_i-q_i|$ from $\sim 0.02$ after FT down to $<0.005$. Residues whose target frequency deviates substantially from the natural-protein average, including the lowest-target residues, are barely moved by FT and are brought close to their target by the RL stage.

\subsection{Composition Alignment Preserves Sequence Quality}
\label{sec:results:quality}
We next ask whether target composition alignment comes at the cost of sequence plausibility. Table~\ref{tab:quality} summarizes three sequence-level quality indicators across all RL runs on both target compositions: predicted solubility (NetSolP), which is relevant because high solubility is a prerequisite for protein digestibility, and two protein-likeness scores, base ProtGPT2 log-PPL (autoregressive, under the original pretrained model) and ESM-2 pPPL (masked-LM, from a different model family).

Mean NetSolP solubility is $0.582\pm 0.030$ on $q_\text{A}$ and $0.628\pm 0.053$ on $q_\text{B}$, both above the NetSolP $0.5$ decision threshold of the original NetSolP study, with individual sequences exceeding $0.9$ on both targets. NetSolP is itself a predictor, so these scores are an \emph{in-silico} sanity check, not a wet-lab claim. The four composition terms are within $0.04$ of each other in mean solubility on $q_\text{A}$ ($0.560$-$0.598$; Table~\ref{tab:rewards}) and within $0.10$ on $q_\text{B}$ ($0.589$-$0.693$; per-variant $q_\text{B}$ values in Table~\ref{tab:rewards-qB}, App.~\ref{app:variants}).

\begin{table*}[ht]
\caption{ESM-2 (150M) pseudo-perplexity (pPPL) as a protein-likeliness measure across pipeline stages, composition term variants, and target compositions; mean$\pm$std over up to $n{=}200$ unique sequences per condition; lower is better. RL rows use the best iteration of one representative seed per variant. ESM-2 pPPL is length-dependent, so we also report mean sequence length $\bar L$.}
\label{tab:esm2}
  \begin{center}
    \begin{small}
      \begin{sc}
        \begin{tabular}{lcccc}
            \toprule
             & \multicolumn{2}{c}{$q_\text{A}$} & \multicolumn{2}{c}{$q_\text{B}$} \\
            \cmidrule(lr){2-3}\cmidrule(lr){4-5}
            Condition & pPPL $\downarrow$ & $\bar L$ & pPPL$\downarrow$ & $\bar L$ \\
            \midrule
            Base ProtGPT2     & 4.48$\pm$1.19 & 387.9 & 4.48$\pm$1.19 & 387.9 \\
            + Domain FT       & 5.74$\pm$2.43 & 375.0 & 6.04$\pm$1.81 & 262.6 \\
            \midrule
            RL cosine             & 10.06$\pm$0.42 & 393.0 & 6.72$\pm$0.33 & 478.9 \\
            RL differentiated     & 9.41$\pm$1.88 & 326.2 & 12.28$\pm$1.19 & 397.0 \\
            RL global-dev.        & 15.79$\pm$1.67 & 480.8 & 7.25$\pm$0.22 & 447.6 \\
            RL symmetric          & 16.65$\pm$0.64 & 495.0 & 10.98$\pm$1.39 & 470.6 \\
            \bottomrule
        \end{tabular}
      \end{sc}
    \end{small}
  \end{center}
  \vskip -0.1in
\end{table*}

\begin{table*}[hbt!]
\caption{Sequence-quality summary, mean$\pm$std across all RL runs per target. NetSolP solubility is a predicted score in $[0,1]$ (higher is more soluble). \emph{base log-PPL} is the mean log-perplexity of the base ProtGPT2 policy. ESM-2 (150M) pseudo-perplexity (pPPL) is computed on up to $n{=}200$ unique sequences per RL composition variant; the entry summarizes mean$\pm$std across the four primary variants. The \emph{iid floor} column reports the same three metrics on $n{=}100$ random sequences sampled iid from $q_\text{A}$ at $L{=}400$, as a composition-only baseline.}
\label{tab:quality}
  \begin{center}
    \begin{small}
      \begin{sc}
        \begin{tabular}{lccc}
            \toprule
            Metric & $q_\text{A}$ ($n{=}120$) & $q_\text{B}$ ($n{=}40$) & iid floor ($q_\text{A}$) \\
            \midrule
            NetSolP solubility    $\uparrow$        & $0.582\pm 0.030$ & $0.628\pm 0.053$ & $0.492\pm 0.055$ \\
            \emph{base log-PPL} $\downarrow$& $9.08\pm 1.13$ & $5.97\pm 0.96$ & $9.22\pm 0.24$ \\
            ESM-2 pPPL $\downarrow$& $12.98\pm 3.77$ & $9.31\pm 2.74$ & $18.07\pm0.57$ \\
            \bottomrule
        \end{tabular}
      \end{sc}
    \end{small}
  \end{center}
  \vskip -0.1in
\end{table*}

ESM-2 pPPL increases monotonically from base ProtGPT2 (mean $4.48$ on both targets, as expected for an unconditional baseline) to the domain-adapted FT prior ($5.74$-$6.04$) to the RL policies ($6.72$-$16.65$, target- and composition-dependent) (Table \ref{tab:esm2}). For reference, ESM-2 pPPL on natural sequences typically falls in the single-digit range~\citep{lin2023esm2}, so the RL policies sit one tier above, indicating a measurable but moderate plausibility cost. This cost is not uniform across composition terms: the \emph{symmetric} composition term in particular almost triples ESM-2 pPPL relative to the FT prior on $q_\text{A}$, which we interpret as the cost of steering the policy toward a non-natural composition without the biological weighting of the \emph{differentiated} term. Notably, the \emph{differentiated} term (mean ESM-2 pPPL $9.41$ on $q_\text{A}$) incurs the smallest increase among the composition variants, suggesting that its biological weighting partially mitigates this plausibility cost. To stress-test the plausibility of the most composition-aligned sequences, we additionally score the top-$30$ sequences from each policy ranked by composition score (i.e.\ the most aggressive on-target sequences each policy generates) (Table~\ref{tab:topk} in App.~\ref{app:topk}). Even on this worst-case slice, no policy collapses to a degenerate high-ESM-2 pPPL regime, and the \emph{differentiated} variant retains its small-fluency-penalty advantage over the other smooth-max variant on $q_\text{A}$.

As a composition-only baseline, we score $n{=}100$ random sequences whose residues are sampled independently and identically distributed (iid) from $q_\text{A}$ at $L{=}400$ on all three metrics (Table~\ref{tab:quality}). Any policy scoring close to this iid floor would be statistically indistinguishable from a random composition-matched sequence in terms of solubility and protein-likeness. The iid floor on ESM-2 pPPL is $18.07$ (approximately length-independent in the relevant range; App.~\ref{app:iid-length}), the iid solubility floor is $0.492\pm 0.055$, and the iid \emph{base log-PPL} floor is $9.22\pm 0.24$. Three observations follow. First, predicted solubility on RL policies ($0.582$ on $q_\text{A}$) sits about $1.6$ floor-std above the iid floor ($0.492$). The gap is modest but consistent in sign, so policy solubility is not entirely a composition artefact. Second, the RL aggregate on $q_\text{A}$ under \emph{base log-PPL} ($9.08\pm 1.13$) is just below the iid floor ($9.22\pm 0.24$); per-variant, \emph{differentiated} ($8.82$) and \emph{cosine} ($7.35$) sit clearly below the floor, while \emph{global-deviation} ($9.98$) and \emph{symmetric} ($10.18$) cross above it. Third, all RL policies remain measurably below the ESM-2 pPPL floor on $q_\text{A}$, so even the worst variant has not collapsed to "random AA'' plausibility. The ESM-2 pPPL gap is, however, variant-dependent: the \emph{differentiated} term ($9.41$) sits roughly midway between the FT prior and the iid floor (about $1.6\times$ the FT pPPL and about half the floor), whereas the \emph{symmetric} term ($16.65$) and \emph{global-deviation} ($15.79$) sit close to the floor (within $\sim 8$-$13\%$ of $18.07$) (Table~\ref{tab:esm2}). This quantifies the qualitative claim above, that $L_p$- and \emph{symmetric} composition terms approach the regime of pure-composition random sequences under ESM-2, while the biologically weighted \emph{differentiated} term retains a clear plausibility margin over that floor.

As a diagnostic on the KL anchor of Eq.~\ref{eq:loss}, we report the \emph{reference log-PPL} (policy mean log-perplexity under its own FT prior), a one-sided proxy for the KL term (differs from $\mathrm{KL}(\pi_\theta\|\pi_\text{ref})$ by the policy entropy). The RL aggregate is $8.30\pm 1.35$ on $q_\text{A}$ and $4.83\pm 0.72$ on $q_\text{B}$, at or below the iid references under the corresponding FT priors ($8.52\pm 0.21$ and $8.90\pm 0.17$; App.~\ref{app:variants}), so no runaway drift on either target. The larger margin on $q_\text{B}$ is consistent with it being the easier target (no zero-target residues). The overlap with the iid reference is expected rather than informative about plausibility, since the FT prior was trained on composition-filtered data, that the same iid-$q_\text{A}$ sequences score $9.22\pm 0.24$ under base ProtGPT2 (Table~\ref{tab:quality}) vs.\ $8.52$ under the FT model, confirming the FT prior has internalized the composition and is a meaningful KL anchor. Target-independent plausibility is read from base log-PPL and ESM-2 pPPL.

Finally, we verify that the RL stage preserves sequence diversity. Table~\ref{tab:diversity-anchors} reports two complementary intra-pool similarity metrics aggregated over all RL runs (lower$=$more diverse). First, \emph{Aln.\ id.}, the mean MMseqs2 identity over alignment-survivors at coverage $\geq 0.6$, which is alignment-conditional and therefore upward-biased toward the most-similar pair tail, and second, \emph{$4$-mer Jacc.}, the mean pairwise Jaccard of $4$-mer sets on a fixed $200$-sequence subsample, which is defined for every pair and so reads the whole pool. Two anchors interpret the magnitudes (Table~\ref{tab:diversity-anchors}, $q_\text{A}\,/\,q_\text{B}$ stacked per cell). Pre-RL pools of $500$ sequences from the FT prior reach \emph{$4$-mer Jacc}.\  $\sim 0.01$ on both targets, which we read as the unconditioned-generation floor at $L\!\sim\!300$ AAs. The RL aggregate ($0.246$ on $q_\text{A}$, $0.483$ on $q_\text{B}$) is measurably more concentrated than this floor, as expected for a policy optimized toward a fixed target composition, but stays well clear of the collapsed RL-only seed of \S\ref{sec:results:ablation:noft} ($0.74$ \emph{$4$-mer Jacc.}, $0.97$ \emph{Aln.\ id.}). Per-variant breakdown is in App.~\ref{app:variants} (Table~\ref{tab:diversity}), and the diversity-pulse mechanism that maintains the margin during training is detailed in App.~\ref{app:divboost} (Fig.~\ref{fig:trajectories}). Mean sequence length stays inside the target interval $[70, 400]$ AA.

\subsection{Ablation Studies}
\label{sec:results:ablation}

We report three controlled ablations: the contribution of the \emph{differentiated} composition term's design choices relative to ablated and baseline alternatives (\emph{symmetric}, \emph{cosine}, and \emph{global-deviation}) on $q_\text{A}$ (\S\ref{sec:results:ablation:reward}), the necessity of the FT stage on $q_\text{A}$ (\S\ref{sec:results:ablation:noft}), and best-of-N selection from the FT prior on both targets (\S\ref{sec:results:ablation:bestofn}).

\begin{table*}[hbt!]
\caption{Intra-pool sequence similarity (lower$=$more diverse), aggregated over all RL runs per target ($n{=}120$ on $q_\text{A}$, $n{=}40$ on $q_\text{B}$; mean$\pm$std), and on two pre-RL anchor pools of $500$ sequences each (base ProtGPT2 and the domain-adaptive FT prior, $q_\text{A}\,/\,q_\text{B}$ stacked per cell). \emph{Aln.\ id.}\ is mean MMseqs2 identity over alignment-survivors at coverage $\geq 0.6$ (alignment-conditional, upward-biased; for the pre-RL anchors, the average is over only $28$-$610$ surviving pairs because most pairs do not align at this threshold, so cross-row comparison should rely on $4$-mer Jacc.). \emph{$4$-mer Jacc.}\ is the mean pairwise Jaccard of $4$-mer sets on a fixed $200$-sequence subsample (defined for every pair, hence directly comparable across rows).}
\label{tab:diversity-anchors}
  \begin{center}
    \begin{small}
      \begin{sc}
        \begin{tabular}{lcccc}
            \toprule
             & \multicolumn{2}{c}{RL policies} & \multicolumn{2}{c}{Pre-RL anchors ($n{=}500$)} \\
            \cmidrule(lr){2-3} \cmidrule(lr){4-5}
            Metric & $q_\text{A}$ & $q_\text{B}$ & FT prior & Base ProtGPT2 \\
            \midrule
            Aln.\ id.$\downarrow$       & $0.832\pm 0.121$ & $0.842\pm 0.212$ & $0.44\,/\,0.54$  & $0.36\,/\,0.36$ \\
            $4$-mer Jacc.$\downarrow$   & $0.246\pm 0.197$ & $0.483\pm 0.345$ & $0.007\,/\,0.010$ & $0.011\,/\,0.010$ \\
            \bottomrule
        \end{tabular}
      \end{sc}
    \end{small}
  \end{center}
  \vskip -0.1in
\end{table*}

\begin{table}[t]
\caption{FT$\rightarrow$RL vs RL-only on $q_\text{A}$, mean across $n{=}3$ paired seeds (same recipe and budget). Comp.\ is the composition score at $\beta_\text{ref}{=}20$; Ess.\ and Pools are sequence constraint indicators (\S\ref{sec:exp}).}
\label{tab:noft}
  \begin{center}
    \begin{small}
      \begin{sc}
        \begin{tabular}{lccccc}
          \toprule
          Setting & Comp.$\uparrow$ & JSD$\downarrow$ & Ess.$\uparrow$ & Pools$\uparrow$ & $N_{\pm 30}\uparrow$ \\
          \midrule
          RL only & 0.420 & 0.032 & 0.24 & 0.46 & 11.9 \\
          FT$\rightarrow$RL       & \textbf{0.574} & \textbf{0.018} & \textbf{0.64} & \textbf{0.87} & \textbf{13.3} \\
          \bottomrule
        \end{tabular}
      \end{sc}
    \end{small}
  \end{center}
  \vskip -0.1in
\end{table}

\begin{table*}[t]
\caption{Reward-formulation comparison on $q_\text{A}$ (mean$\pm$std across $n=30$ seeds, matched compute). Solub.\ is mean NetSolP predicted solubility (higher$=$more soluble). Best per column in bold.}
\label{tab:rewards}
  \begin{center}
    \begin{small}
      \begin{sc}
        \begin{tabular}{lccccc}
            \toprule
            Reward & Comp.$\uparrow$ & JSD$\downarrow$ & $N_{\pm 30}$$\uparrow$ & $L_1$$\downarrow$ & Solub.$\uparrow$ \\
            \midrule
            differentiated & \textbf{0.397$\pm$0.210} & 0.032$\pm$0.020 & 12.34$\pm$2.57 & 0.259$\pm$0.091 & \textbf{0.598$\pm$0.031} \\
            symmetric & 0.305$\pm$0.167 & \textbf{0.028$\pm$0.012} & \textbf{12.89$\pm$2.02} & \textbf{0.247$\pm$0.069} & 0.582$\pm$0.027  \\
            cosine & 0.280$\pm$0.188 & 0.033$\pm$0.013 & 12.51$\pm$2.53 & 0.258$\pm$0.091 & 0.586$\pm$0.032 \\
            global-deviation & 0.140$\pm$0.028 & 0.041$\pm$0.003 & 11.44$\pm$0.49 & 0.312$\pm$0.013 & 0.560$\pm$0.010 \\
            \bottomrule
        \end{tabular}
      \end{sc}
    \end{small}
  \end{center}
  \vskip -0.1in
\end{table*}

\subsubsection{Composition term formulation}
\label{sec:results:ablation:reward}

We evaluate the design choices of the \emph{differentiated} composition term against one ablated (\emph{symmetric}) and two baseline (\emph{cosine} and  \emph{global-deviation}) composition terms on $q_\text{A}$ at matched compute and shared seeds (Table~\ref{tab:rewards}). Pairwise Wilcoxon signed-rank tests on the $n{=}30$ paired seeds give a clear ranking: every smooth-max (\emph{differentiated}, \emph{symmetric}) or \emph{cosine} term beats the $L_1$ term (\emph{global-deviation}) on composition score (all surviving Bonferroni correction at $\alpha=0.05$), and on JSD for \emph{symmetric} over \emph{global-deviation}. The remaining three variants (differentiated, symmetric, cosine) are not pairwise separable on JSD, and only \emph{differentiated} vs.\ \emph{cosine} separates uncorrected on the composition score. Full pairwise statistics are in App.~\ref{app:wilcoxon-cis}; per-variant seed distributions in App.~\ref{app:variants}.

On this target, the statistical comparison confirms that any smooth-max- or cosine-based composition term substantially outperforms the $L_1$ baseline, while the three upper-cluster variants are not reliably separable on aggregate metrics alone. However, the \emph{differentiated} term is our default choice for two reasons that go beyond aggregate composition metrics alone. First, it directly encodes biological constraints, penalizing deficits in essential AA more harshly than excesses, handling interchangeable AA pools as group-level terms, and amplifying deviations on zero-target AA, which are design choices motivated by the nutritional objective rather than arbitrary hyperparameters. Second, it incurs the smallest pPPL penalty among the smooth-max variants (\S\ref{sec:results:quality}, Table~\ref{tab:esm2}), suggesting that biological weighting partially mitigates the plausibility cost of composition alignment.

\subsubsection{Domain-adaptive FT vs RL-only}
\label{sec:results:ablation:noft}

We assess the necessity of the domain-adaptive FT stage by running RL directly on base ProtGPT2 (RL-only), on $n{=}3$ seeds shared with the FT $\to$ RL runs (Table~\ref{tab:noft}). Mean paired differences (FT$\to$RL minus RL-only across the three seeds) favor FT $\to$ RL on every metric: composition $+0.15$, essential-residue coverage $+0.40$, pool tolerance $+0.41$, $N_{\pm 30}{+}1.4$, and JSD $-0.014$ (lower is better). Given only $n{=}3$ paired seeds, we report direction and magnitude without formal significance testing. The FT$\to$RL values of this subsection are restricted to the three seeds matched to the RL-only version, and it is therefore not directly comparable to the $n{=}30$ mean reported in \S\ref{sec:results:rl}. Per-seed inspection reveals a characteristic reward-hacking failure mode that emerges when RL is run without the FT prior, that is, one of the three RL-only seeds collapsed onto a very narrow AA palette, satisfying low-target compliance perfectly (no zero-target AA ever exceeded its cap), but in doing so, it dropped essential-residue coverage to zero. Despite this, it still scored a high composition score of $0.70$ because the average frequency vector remained close to $q_\text{A}$. 

The collapse is also visible as a diversity signature. The failed RL-only seed reaches a mean intra-pool pairwise identity of $0.97$ (near saturation; \emph{Aln.\ id.}) and \emph{$4$-mer Jaccard}  of $0.74$, versus $0.84$/$0.89$ \emph{Aln.\ id.}\ and $0.21$/$0.31$ \emph{$4$-mer Jacc}.\  for the surviving RL-only seeds, and $0.72$-$0.82$ \emph{Aln.\ id.}\ ($0.52$-$0.57$ $4$-mer Jacc.) for the matched FT$\to$RL seeds, suggesting on both metrics that, without the FT prior, the policy contracts onto a narrow palette of nearly-identical sequences. This single seed inflates the RL-only mean composition score in Table~\ref{tab:noft}, but the other two RL-only seeds average composition $\approx 0.29$. FT prevents this failure mode in our sweep, though we note this is based on a single RL-only failure case and a limited $n=3$ paired comparison. We read the mechanism as a reward-landscape effect rather than a property of the FT prior in isolation. The FT prior is itself narrower than base ProtGPT2 (its training set is composition-filtered), but it sits much closer to $q$ ($\mathrm{JSD}\,0.059$ vs.\ $0.247$), so reward-weighted updates are spread across many moderately-rewarding directions and the KL penalty against $\pi_\text{ref}$ (\S\ref{sec:method:rl}) actively pulls the policy back toward this broad composition-conditioned region. From base ProtGPT2, the same KL budget cannot reach a comparably-rewarded region without latching onto a few high-reward modes, which is the collapse signature observed here.


\subsubsection{Best-of-N from the FT prior}
\label{sec:results:ablation:bestofn}

To test whether reward-weighted RL updates merely reproduce best-of-N selection from the FT prior, we draw $500$ sequences from the FT-only model and from base ProtGPT2, and rank them by the same fixed-$\beta_\text{ref}$ composition score used throughout. On $q_\text{A}$, the best-of-$500$ from FT-only reaches $\mathrm{JSD}=0.018$ (composition score$=0.529$), while the best FT$\to$RL run reaches $\mathrm{JSD}=0.0044$ (composition score $=0.822$). This is a $4\times$ reduction in JSD that no amount of selection from the FT prior recovers in $500$ draws. On $q_\text{B}$, best-of-$500$ from FT-only reaches $\mathrm{JSD}=0.0111$ vs.\ $0.0008$ for the best FT$\to$RL run, a $14\times$ gap. Base ProtGPT2 is two orders of magnitude further from either target even at the best-of-$500$ tail (Table~\ref{tab:bon} in App.~\ref{app:bestofn}). The RL stage, therefore, moves the policy into a region of sequence space that the FT prior does not reach by oversampling alone.


\section{Discussion}
\label{sec:disc}

This work demonstrates that steering a PLM toward an explicit distributional target AA composition benefits from two training stages addressing different aspects of the problem. Domain-adaptive FT shifts the base model toward an average composition close to $q$, but cannot enforce the sequence constraints needed to select individual candidates for synthesis. This per-residue gap is largest on residues whose target frequency is far from the natural protein average. The RL stage closes this gap, taking the FT prior from weak sequence constraint satisfaction to near-saturation under the best aligned policy, a transition that oversampling from the FT prior alone cannot reproduce. Removing the FT stage degrades every metric on every paired seed, and one of the three RL-only seeds exhibits a discrete failure mode in which the policy reward-hacks by collapsing onto a narrow residue palette that satisfies the target composition on average while suppressing essential residue coverage. This failure mode is absent from all FT$\to$RL runs in our sweep, supporting the view that the domain-adaptive FT provides a stable initialization that the reward signal alone cannot guarantee.

A key contribution of this work is the \emph{differentiated} composition term, a novel reward formulation tailored to the specific task of designing digestible proteins with a nutrition-oriented AA composition. To evaluate its design choices, we compare it against an ablated variant (\emph{symmetric}) and two existing baselines (\emph{cosine}, \emph{global-deviation}) in a controlled seed-matched experiment. The results show that the \emph{global-deviation} baseline is clearly insufficient while the remaining three terms form a broadly comparable upper cluster. A weak signal suggests that smooth-max variants may outperform the \emph{cosine} term, though this does not survive Bonferroni correction at the seed counts used here. Within the upper cluster, the \emph{differentiated} term attains the highest mean composition score on the $q_A$ target composition while its best run saturates all sequence constraints (essential-residue coverage, pool tolerance, low-target compliance), making it the most faithful to the nutritional design objective. Additionally, \emph{differentiated} carries the smallest ESM-2 pPPL penalty, indicating that the composition gains are not bought at a disproportionate plausibility cost, and suggesting that biologically grounded reward design may inherently produce more natural-looking sequences than uniform alternatives.

Several limitations should be noted. First, the pipeline does not enforce sequence plausibility beyond the regularization provided by the KL penalty against the FT reference model, and we do not measure structure or function directly. Second, seed counts are modest ($30$ on $q_\text{A}$ and $10$ on $q_\text{B}$), so we report bootstrap CIs rather than asymptotic null tests. Third, biological follow-up, including digestibility-related proxies, is necessary to determine whether improved target composition alignment also yields biologically plausible candidates. Addressing these limitations represents the natural next step for this line of work.

To the best of our knowledge, this is the first work to steer a PLM toward an explicit target AA composition as a primary design objective. While matching a frequency vector over 20 AA may appear to be a simple distributional matching problem, it is better understood as a multi-objective sequence design task: generated sequences must simultaneously satisfy a range length, essential-residue coverage, interchangeable pool balance, and zero-target compliance, without drifting away from the manifold of plausible protein sequences. That the best aligned policies satisfy all these constraints simultaneously while retaining reasonable ESM-2 plausibility demonstrates that the proposed two-stage pipeline, domain-adaptive FT followed by reward-weighted RL with a biologically grounded composition term, is a viable approach to this class of constrained generative objectives.


\section*{Impact Statement}

This work develops a method for steering PLMs toward an explicit target AA composition. The motivating application is feed-protein design, but the pipeline is broadly applicable to any composition-constrained protein generation task. All generated sequences are computational proposals that require further validation before wet-lab synthesis or downstream deployment. Depending on the application, this includes structure prediction, disorder analysis, digestibility assays for feed proteins, or activity assays for functional proteins. Potential positive impacts include better-aligned dietary proteins for animal feed and a reduced environmental cost of feed production.

\section*{Acknowledgements}
This work was supported by the European Innovation Council (EIC) Pathfinder Open project SYNFEED, under grant agreement No. 101186580. The European Commission's support for the production of this publication does not constitute endorsement of the contents, which reflects the views only of the authors, and the Commission cannot be held responsible for any use which may be made of the information contained therein.

\bibliography{example_paper}
\bibliographystyle{icml2026}

\newpage
\appendix
\onecolumn
\section{Reproducibility details}\label{app:repro}

This appendix consolidates the practical settings used to produce every number in the main text. All runs used a single NVIDIA H100 GPU and a single shared conda environment.



\textbf{Dataset construction.} Filters and final dataset sizes (\S\ref{sec:method:ft}): UniProtKB/TrEMBL~\citep{uniprot2023} (FASTA from the UniProt Consortium FTP) $\to$ length filter $[100, 500]$ aa $\to$ per-sequence cosine similarity to $q$ $\geq 0.95$ $\to$ MMseqs2 redundancy at $<70\%$ pairwise identity, yielding $\sim$$1.0{\times}10^{5}$ sequences split 50/50 into train and eval per target.

\textbf{Stage 1 (FT).} ProtGPT2 base; ProtGPT2 tokenizer; block size $512$; AdamW with learning rate $5{\times}10^{-5}$, weight decay $0.01$, $3\%$ linear warm-up; effective batch size $32$ blocks; up to $40$ epochs with early stopping on eval loss (patience $7$, $\min\Delta=10^{-3}$). The selected checkpoint for $q_\text{A}$ is at epoch $21$; total wall-clock $\sim$$13.5$\,h. The same checkpoint is reused as $\pi_\text{ref}$ for every RL run on $q_\text{A}$ while  $q_\text{B}$ uses its own FT checkpoint trained with identical hyperparameters on the $q_\text{B}$ filtered dataset.

\textbf{Stage 2 (RL).} Settings shared across all reported runs (Table~\ref{tab:hparams}).

\begin{table}[h]
\caption{Stage-2 (RL) hyperparameters. Values are fixed across all $160$ runs reported in this paper.}
\label{tab:hparams}
  \begin{center}
    \begin{small}
      \begin{sc}
        \begin{tabular}{ll}
        \toprule
        Setting & Value \\
        \midrule
        Outer iterations         & $50$ \\
        Candidates per iteration & $2000$ \\
        Diversity filter         & MMseqs2, $<0.85$ pairwise identity \\
        Length filter            & target interval, outliers dropped \\
        Reward weights $(w_c,w_l)$ & $(0.97, 0.03)$ static \\
        Sharpness coefficient $\beta$ & ramp $15{\to}36$ during training \\
        \quad evaluation $\beta_\text{ref}$ & $20$ (fixed) \\
        Policy LR                & $3{\times}10^{-6}$ \\
        Optimizer                & AdamW, default betas \\
        KL weight $\lambda_\text{KL}$ & $0.05\to 0.15$ over first $55\%$ \\
        Reference model          & FT checkpoint, frozen \\
        \bottomrule
        \end{tabular}
        \end{sc}
        \end{small}
        \end{center}
\end{table}

\textbf{Run breakdown.} The full $4{\times}30 + 4{\times}10 = 160$ runs of \S\ref{sec:exp:runs} share the loop and optimizer settings of Table~\ref{tab:hparams}. The only differences between runs are the random seed, the composition variant, and the target.

\textbf{Evaluation protocol.} For each run, we select the iteration that maximizes the fixed $\beta_\text{ref}{=}20$ composition score on its candidate pool, and report all metrics on that pool. Per-variant confidence intervals are seed-level $95\%$ percentile bootstraps with $10^4$ resamples.

\section{Composition term formulas}\label{app:rewards}

We give explicit formulas for all composition terms used at training and evaluation time. Let $p\in\mathcal{S}_{20}$ be the observed AA frequency vector of a candidate sequence (over the 20 standard AA) and $q\in\mathcal{S}_{20}$ the target frequency vector. Let $L$ denote sequence length and let $\mathcal{E}$ denote the essential-AA index set used in the \emph{differentiated} composition.

\textbf{Per-residue error (\emph{differentiated}).} For each AA $i$,
\begin{equation*}
e_i =
\begin{cases}
w^- (q_i - p_i)               & i\in\mathcal{E},\ p_i < q_i,\\
w^+ (p_i - q_i)               & i\in\mathcal{E},\ p_i \geq q_i,\\
w^{\text{ne}}\,|p_i - q_i|    & i\notin\mathcal{E},
\end{cases}
\end{equation*}
with $w^- = 3.0$, $w^+ = 0.35$, $w^{\text{ne}} = 1.0$, plus a zero-target amplifier $e_i \leftarrow \alpha_0\, e_i$ whenever $q_i = 0$ and $p_i > 0$ (with $\alpha_0 = 3$). Per-residue errors are clipped to $\le 2$. Group-level errors $w^{\text{grp}}\,|\sum_{i\in G} p_i - t_G|$ for interchangeable AA groups $G$ (with $w^{\text{grp}}=1.5$; on $q_\text{A}$ these are the sulfur and aromatic-precursor pools, \S\ref{sec:exp}) are appended to the error vector $e$.

\textbf{\emph{Differentiated} score.} With $\alpha = 60$, $w_\text{rms} = 1$, $w_\text{sm} = 0.9$, and $\beta$ ramped $15{\to}36$ across training iterations, and letting $\mathrm{rms}(e) = \sqrt{\tfrac{1}{n}\sum_i e_i^2}$ and $\mathrm{smax}_\alpha(e) = \tfrac{1} {\alpha}\log\!\big(\tfrac{1}{n}\sum_i e^{\alpha e_i}\big)$,
\begin{equation*}
\mathrm{Comp}_\text{diff}(p; q) =
\exp\!\Big(\!-\beta\big[w_\text{rms}\,\mathrm{rms}(e) + w_\text{sm}\,\mathrm{smax}_\alpha(e)\big]\Big).
\end{equation*}
The second bracket term is a smooth-max (log-sum-exp) over the per-residue errors with sharpness $\alpha$. Evaluation uses $\beta_\text{ref}=20$.

\textbf{\emph{Symmetric} score.} Same RMS + smooth-max kernel, but with uniform absolute deviation $e_i = |p_i - q_i|$ (no essential split, no group terms). The remaining hyperparameters $(\alpha, w_\text{rms}, w_\text{sm}, \alpha_0)$ are kept identical to the \emph{differentiated} variant; only the outer sharpness $\beta$ is ramped on a different schedule, $30{\to}80$ across training iterations, to match the loss scale of the simpler kernel:
\begin{equation*}
\mathrm{Comp}_\text{sym}(p; q) =
\exp\!\Big(\!-\beta\big[w_\text{rms}\,\mathrm{rms}(e) + w_\text{sm}\,\mathrm{smax}_\alpha(e)\big]\Big).
\end{equation*}
\emph{differentiated} and \emph{symmetric} therefore share the same $\mathrm{RMS}+\mathrm{smooth\text{-}max}$ kernel; the \emph{differentiated} variant adds biologically-motivated essential/non-essential weighting and the interchangeable-group terms, while \emph{symmetric} uses uniform $|p_i-q_i|$ throughout.

\textbf{\emph{Cosine} score.}
\begin{equation*}
\mathrm{Comp}_\text{cos}(p; q) =
\exp\!\Big(-\beta\big[1 - \tfrac{p\cdot q}{\|p\|\,\|q\|}\big]\Big),
\end{equation*}
with $\beta$ ramped $5{\to}25$ across training iterations. Scale-invariant: matches direction in $\mathbb{R}^{20}$, not magnitude, which is why it lags on the $L_1$ metric.

\textbf{\emph{Global-deviation} score.}
\begin{equation*}
\mathrm{Comp}_\text{gd}(p; q) = \exp\!\Big(-\beta\,\tfrac{1}{20}\textstyle\sum_i |p_i - q_i|\Big),
\end{equation*}
with $\beta$ ramped $20{\to}100$ across iterations. The simplest baseline: mean-$L_1$ deviation, no smooth-max, no essential split.

\textbf{Per-variant $\beta$ ramps.} The four ramps above ($15{\to}36$, $30{\to}80$, $5{\to}25$, $20{\to}100$) were chosen based on the best results that were achieved. Within each variant, $\beta$ is linearly ramped over a fixed window of iterations (starting at iteration $9$, target value reached by iteration $33$, then held constant).

\textbf{Reference-$\beta$ re-scoring.} For all aggregate comparisons, we rescore each candidate pool with the \emph{differentiated} formula at $\beta_\text{ref}{=}20$, regardless of which variant was used during training. This puts every variant on a single common scoring function and removes confounding from the per-variant $\beta$ schedule.

\textbf{Length term.} A piecewise-linear shaping term over the support interval $[L_\text{min},L_\text{max}]$ with peak plateau $[L_a,L_b]$:
\begin{equation*}
R_\ell(L) =
\begin{cases}
0 & L \leq L_\text{min}\ \text{or}\ L \geq L_\text{max},\\
\dfrac{L - L_\text{min}}{L_a - L_\text{min}} & L_\text{min} < L < L_a,\\[2pt]
1 & L_a \leq L \leq L_b,\\[2pt]
\dfrac{L_\text{max} - L}{L_\text{max} - L_b} & L_b < L < L_\text{max},
\end{cases}
\end{equation*}
with $L_\text{min}=70$, $L_a=110$, $L_b=250$, $L_\text{max}=400$ AA. The non-zero support $[70, 400]$ AA is intentionally wider than the length filter $[100, 500]$ AA (\S\ref{sec:method:ft}) so that on-target sequences just outside the dataset window still receive partial credit.

\section{Top-5 runs}\label{app:top5}
Table~\ref{tab:top5jsd} lists the top RL runs on $q_\text{A}$ across all $n{=}120$ runs, with rank columns for both JSD (lower better) and fixed-$\beta_\text{ref}$ composition score (higher better) so that disagreements between the two rankings are visible.

\begin{table}[h!]
\caption{Top-5 RL runs on $q_\text{A}$ (across all $n{=}120$
runs) ranked by JSD (lower is better).}
\label{tab:top5jsd}
  \begin{center}
    \begin{small}
      \begin{sc}
        \begin{tabular}{ccllcccccc}
        \toprule
        Rank-JSD & Rank-Comp & Reward & Comp.$\uparrow$ & JSD$\downarrow$ & $N_{\pm 30}\uparrow$ & Essent.$\uparrow$ & Pools$\uparrow$ & Low-tgt.$\uparrow$ \\
        \midrule
        1 & 1 & differentiated & 0.822 & 0.0044 & 17.37 & 1.000 & 1.000 & 1.000 \\
        2 & -- & symmetric      & 0.618 & 0.0049 & 19.00 & 1.000 & 1.000 & 1.000 \\
        3 & 3 & symmetric      & 0.761 & 0.0049 & 17.03 & 1.000 & 1.000 & 1.000 \\
        4 & 2 & cosine         & 0.797 & 0.0060 & 17.00 & 1.000 & 1.000 & 1.000 \\
        5 & 5 & differentiated & 0.669 & 0.0073 & 16.75 & 0.588 & 0.591 & 0.977 \\
        -- & 4 & differentiated & 0.690 & 0.0086 & 16.38 & 1.000 & 1.000 & 1.000 \\
        \bottomrule
        \end{tabular}
        \end{sc}
    \end{small}
  \end{center}
\end{table}

\section{Composition-variant breakdown}\label{app:variants}

This appendix collects the per-variant figures referenced from
\S\ref{sec:results:ablation:reward}, and the analogous composition-variant comparison on $q_\text{B}$ ($n{=}10$ seeds per variant; Table~\ref{tab:rewards-qB}).

\begin{table}[h]
\caption{Reward-formulation comparison on $q_\text{B}$ (mean$\pm$std across $n{=}10$ seeds, matched compute). Same metrics, scoring, and selection rule as Table~\ref{tab:rewards}. Best per column in bold.}
\label{tab:rewards-qB}
  \begin{center}
    \begin{small}
      \begin{sc}
        \begin{tabular}{lccccc}
            \toprule
            Reward & Comp.$\uparrow$ & JSD$\downarrow$ & $N_{\pm 30}$$\uparrow$ & $L_1$$\downarrow$ & Solub.$\uparrow$ \\
            \midrule
            differentiated   & 0.566$\pm$0.117 & 0.021$\pm$0.007 & 12.21$\pm$2.18 & 0.257$\pm$0.065 & 0.589$\pm$0.036 \\
            symmetric        & \textbf{0.672$\pm$0.165} & \textbf{0.009$\pm$0.008} & \textbf{17.11$\pm$2.31} & \textbf{0.117$\pm$0.071} & \textbf{0.693$\pm$0.019} \\
            cosine           & 0.524$\pm$0.171 & 0.016$\pm$0.010 & 14.02$\pm$3.12 & 0.204$\pm$0.090 & 0.628$\pm$0.024 \\
            global-deviation & 0.460$\pm$0.131 & 0.023$\pm$0.009 & 12.12$\pm$2.86 & 0.265$\pm$0.081 & 0.603$\pm$0.048 \\
            \bottomrule
        \end{tabular}
      \end{sc}
    \end{small}
  \end{center}
  \vskip -0.1in
\end{table}

\begin{figure}[htb!]
  \begin{center}
  \centerline{\includegraphics[width=\columnwidth]{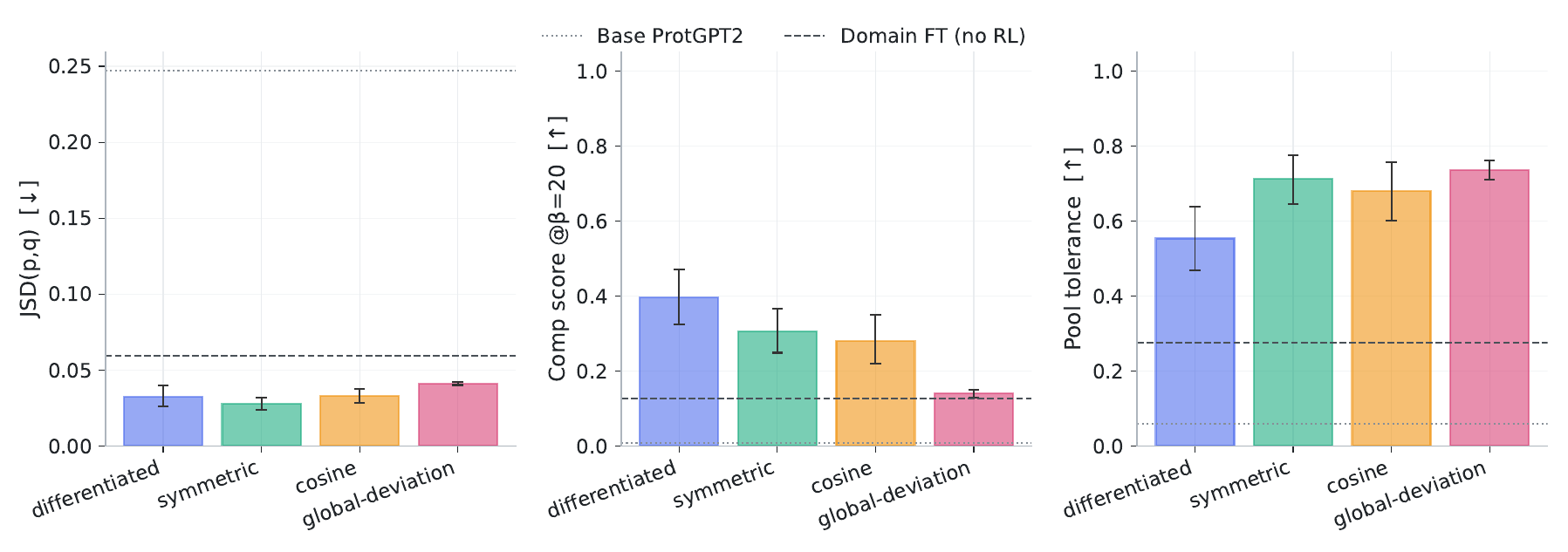}}
\caption{Per-variant means with seed-level $95\%$ bootstrap confidence intervals on JSD, composition score (re-scored at fixed $\beta_\text{ref}{=}20$), and pool tolerance ($q_\text{A}$, $n{=}30$ seeds per variant). Dotted line: base ProtGPT2; dashed line: domain-adaptive FT (no RL).}
\label{fig:panel}
\end{center}
\end{figure}

\begin{figure}[htb!]
   \vskip 0.1in
  \begin{center}
  \centerline{\includegraphics[width=\columnwidth]{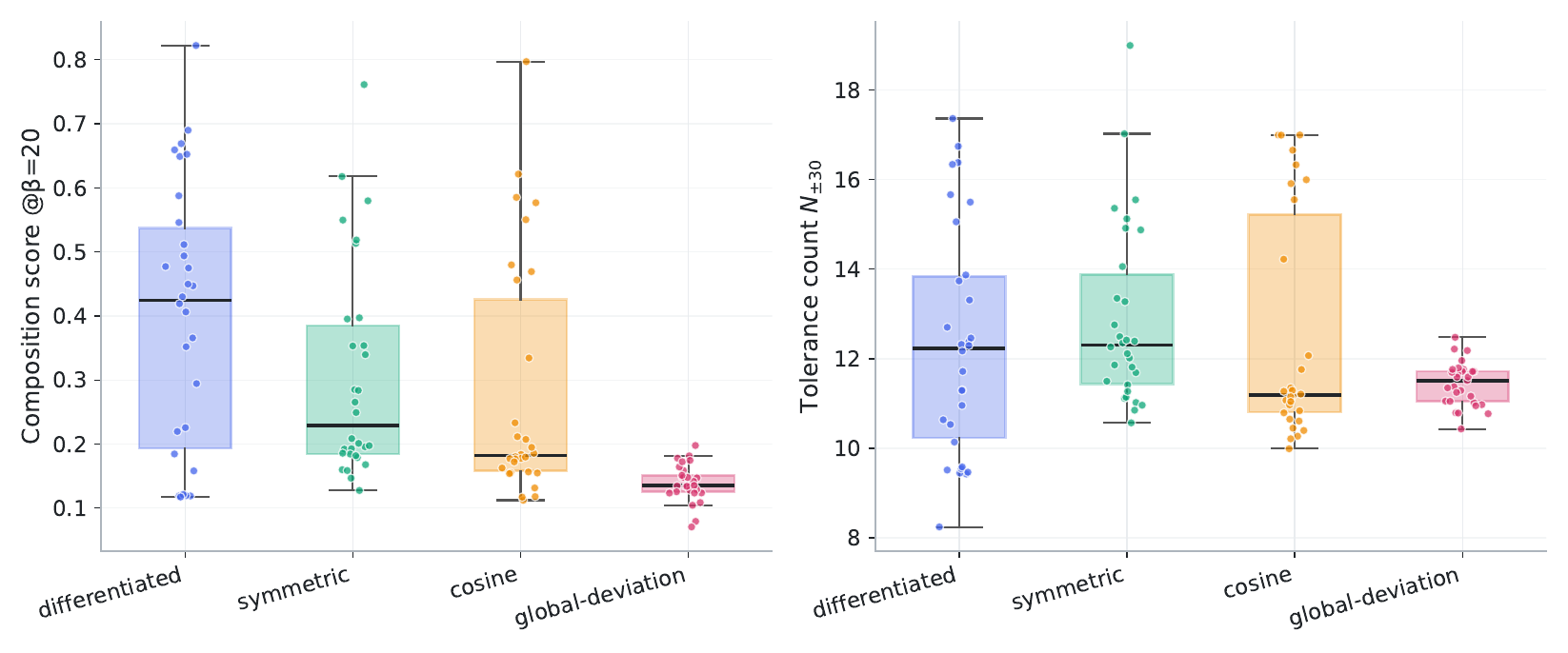}}
\caption{Seed variance per composition term variant on the composition score (left, at $\beta_\text{ref}{=}20$) and the tolerance count $N_{\pm 30}$ (right) on $q_\text{A}$, $n{=}30$ seeds per variant. Boxes summarize the seed distribution; jittered points show individual seeds.}
\label{fig:seedvar}
\end{center}
\end{figure}

\begin{table}[h!]
\caption{Per-variant intra-pool sequence-similarity breakdown on $q_\text{A}$ ($n{=}30$ seeds per variant) and $q_\text{B}$ ($n{=}10$ seeds per variant), mean$\pm$std (lower$=$more diverse). \emph{Aln.\ id.}\ is mean MMseqs2 identity over alignment-survivors at coverage $\geq 0.6$. \emph{$4$-mer Jacc.}\ is the mean pairwise Jaccard over $4$-mer sets on a fixed $200$-sequence subsample.}
\label{tab:diversity}
  \begin{center}
    \begin{small}
      \begin{sc}
        \begin{tabular}{llcc}
            \toprule
            Target & Reward & Aln.\ id.$\downarrow$ & $4$-mer Jacc.$\downarrow$ \\
            \midrule
            $q_\text{A}$ & differentiated   & 0.849$\pm$0.092 & 0.334$\pm$0.244 \\
            $q_\text{A}$ & symmetric        & \textbf{0.837$\pm$0.106} & 0.288$\pm$0.165 \\
            $q_\text{A}$ & cosine           & 0.848$\pm$0.185 & 0.307$\pm$0.271 \\
            $q_\text{A}$ & global-deviation & 0.862$\pm$0.081 & \textbf{0.143$\pm$0.018} \\
            \midrule
            $q_\text{B}$ & differentiated   & 0.873$\pm$0.090 & 0.411$\pm$0.312 \\
            $q_\text{B}$ & symmetric        & 0.855$\pm$0.291 & 0.766$\pm$0.268 \\
            $q_\text{B}$ & cosine           & \textbf{0.794$\pm$0.280} & 0.484$\pm$0.348 \\
            $q_\text{B}$ & global-deviation & 0.847$\pm$0.072 & \textbf{0.272$\pm$0.239} \\
            \bottomrule
        \end{tabular}
      \end{sc}
    \end{small}
  \end{center}
  \vskip -0.1in
\end{table}

\begin{table}[h!]
\caption{Log-perplexity of each RL policy under its FT prior (one-sided proxy for $\mathrm{KL}(\pi_\theta\|\pi_\text{ref})$, equal to $\mathbb{E}_{\pi_\theta}[-\log\pi_\text{ref}]$ up to the policy entropy, \S\ref{sec:results:quality}). Each cell is mean$\pm$std across seeds of per-run pool means. Values are \emph{not} directly comparable between targets because each target uses a different FT prior. The \emph{Aggregate} row pools all four variants per target. The \emph{iid floor} row reports the same metric on $n{=}100$ random sequences sampled iid from the corresponding target at $L{=}400$, scored under that target's FT prior.}

\label{tab:ref-logppl}
  \begin{center}
    \begin{small}
      \begin{sc}
        \begin{tabular}{lcc}
            \toprule
            Condition & $q_\text{A}$ & $q_\text{B}$ \\
            \midrule
            RL cosine             & $8.28\pm 0.63$ ($n{=}30$) & $4.96\pm 0.79$ ($n{=}10$) \\
            RL differentiated     & $8.10\pm 2.35$ ($n{=}30$) & $4.50\pm 0.81$ ($n{=}10$) \\
            RL global-dev.        & $8.36\pm 0.80$ ($n{=}30$) & $4.85\pm 0.43$ ($n{=}10$) \\
            RL symmetric          & $8.47\pm 0.88$ ($n{=}30$) & $5.02\pm 0.76$ ($n{=}10$) \\
            \midrule
            Aggregate             & $8.30\pm 1.35$ ($n{=}120$) & $4.83\pm 0.72$ ($n{=}40$) \\
            \midrule
            iid floor  & $8.52\pm 0.21$ ($n{=}100$) & $8.90\pm 0.17$ ($n{=}100$) \\
            \bottomrule
        \end{tabular}
      \end{sc}
    \end{small}
  \end{center}
  \vskip -0.1in
\end{table}

\subsection{Pairwise Wilcoxon: full statistics}
\label{app:wilcoxon-cis}

Table~\ref{tab:wilcoxon-full} reports the median paired difference $\Delta$, $95\%$ percentile-bootstrap CI, and Wilcoxon $p$-value for every pair and every aggregate metric on $q_\text{A}$.

\begin{table}[h!]
\caption{Full pairwise Wilcoxon signed-rank statistics on
$q_\text{A}$, $n{=}30$ paired seeds. Bold $p$ survives Bonferroni
at $\alpha{=}0.05$ within each metric.}
\label{tab:wilcoxon-full}
  \begin{center}
    \begin{small}
      \begin{sc}
        \begin{tabular}{lllrrl}
        \toprule
        Metric & A & B & $\Delta$ & 95\% CI & $p$ \\
        \midrule
        \multirow{6}{*}{Comp.} 
         & diff & sym & $+0.069$ & $[-0.059, +0.166]$ & 0.10 \\
         & diff & cos & $+0.118$ & $[+0.003, +0.252]$ & 0.039 \\
         & diff & gd  & $+0.271$ & $[+0.111, +0.351]$ & $\mathbf{<10^{-5}}$ \\
         & sym  & cos & $+0.031$ & $[-0.024, +0.173]$ & 0.40 \\
         & sym  & gd  & $+0.114$ & $[+0.050, +0.214]$ & $\mathbf{<10^{-6}}$ \\
         & cos  & gd  & $+0.054$ & $[+0.030, +0.084]$ & $\mathbf{<10^{-4}}$ \\
        \midrule
        \multirow{6}{*}{JSD}
         & diff & sym & $+0.0006$ & $[-0.0059, +0.0172]$ & 0.30 \\
         & diff & cos & $-0.0034$ & $[-0.0125, +0.0107]$ & 0.77 \\
         & diff & gd  & $-0.0130$ & $[-0.0207, +0.0020]$ & 0.017 \\
         & sym  & cos & $-0.0076$ & $[-0.0165, +0.0000]$ & 0.088 \\
         & sym  & gd  & $-0.0120$ & $[-0.0184, -0.0045]$ & $\mathbf{<10^{-6}}$ \\
         & cos  & gd  & $-0.0033$ & $[-0.0069, +0.0008]$ & 0.024 \\
        \midrule
        \multirow{6}{*}{$N_{\pm 30}$}
         & diff & sym & $-0.40$ & $[-1.99, +1.19]$ & 0.30 \\
         & diff & cos & $-0.05$ & $[-1.66, +1.35]$ & 0.75 \\
         & diff & gd  & $+0.99$ & $[-1.13, +1.86]$ & 0.21 \\
         & sym  & cos & $+0.67$ & $[+0.05, +1.70]$ & 0.37 \\
         & sym  & gd  & $+0.86$ & $[+0.30, +1.68]$ & $\mathbf{<10^{-4}}$ \\
         & cos  & gd  & $-0.08$ & $[-0.55, +0.81]$ & 0.40 \\
        \bottomrule
        \end{tabular}
        \end{sc}
    \end{small}
  \end{center}
\end{table}

\section{Best-of-$N$ vs.\ RL}\label{app:bestofn}

Table~\ref{tab:bon} expands the rejection-sampling argument of \S\ref{sec:results:rl}. We score $500$ samples from each prior (base ProtGPT2 and FT-only) on each target with the fixed-$\beta_\text{ref}{=}20$ composition score and report the best and top-$10\%$ JSD alongside the best composition score. The best-of-$500$ from the FT prior is closer to the target than from base ProtGPT2 by a factor of two to three on JSD, confirming that domain-adaptive FT is a real prior shift, but it is still $4\times$ ($q_\text{A}$) to $14\times$ ($q_\text{B}$) further from the target than the best RL run. The gap is largest on the per-sequence composition score, where the FT prior simply does not contain sequences that satisfy the harder per-residue constraints.

\begin{table*}[h!]
\caption{Rejection-sampling baseline. For each target, we draw $N{=}500$ sequences from base ProtGPT2 and from the FT-only model and rank them by the same fixed-$\beta_\text{ref}$ composition score used elsewhere. We report the best and top-$10\%$ JSD alongside the best composition score, compared against the best RL run on each target. Even with $N{=}500$ draws, the FT prior alone does not reach the JSD or score that the RL stage reaches in a single run, on either target. Values for the two ``RL best'' rows are the per-run pool means at the best iteration ($n{=}19$, $n{=}39$ valid sequences after filtering on $q_\text{A}$ and $q_\text{B}$ respectively).}
\label{tab:bon}
  \begin{center}
    \begin{small}
      \begin{sc}
        \begin{tabular}{llrrrr}
            \toprule
            Target & Source & $N$ & Score$_\text{best}\uparrow$ & JSD$_\text{best}\downarrow$ & JSD$_\text{p10}\downarrow$ \\
            \midrule
            $q_\text{A}$ & base ProtGPT2 (B/N) & 500 & 0.146 & 0.041 & 0.091 \\
            $q_\text{A}$ & FT only (B/N)       & 500 & 0.529 & 0.018 & 0.028 \\
            $q_\text{A}$ & RL best run         &  19 & \textbf{0.822} & \textbf{0.004} & --    \\
            \midrule
            $q_\text{B}$ & base ProtGPT2 (B/N) & 500 & 0.446 & 0.024 & 0.076 \\
            $q_\text{B}$ & FT only (B/N)       & 500 & 0.619 & 0.011 & 0.023 \\
            $q_\text{B}$ & RL best run         &  39 & \textbf{0.830} & \textbf{0.001} & --    \\
            \bottomrule
            \end{tabular}
      \end{sc}
    \end{small}
  \end{center}
  \vskip -0.1in
\end{table*}

\section{Top-$30$ ESM-2 pPPL}\label{app:topk}
The per-condition top-$30$ slice referenced in \S\ref{sec:results:quality} is computed by ranking each policy's candidate pool (the same generation pool that backs Table~\ref{tab:esm2}) by the fixed-$\beta_\text{ref}$ composition score and scoring the top $30$ with ESM-2 (150M) pPPL (Table \ref{tab:topk}). The top-$30$ sequences are those most aggressive in matching $q$ from each policy, so this is the worst-case fluency snapshot (rather than the average over the whole pool reported in Table~\ref{tab:esm2}). Even on the per-sequence top-$30$ slice, no variant collapses to a degenerate-low-perplexity regime: all RL policies remain in the same order of magnitude as their respective FT prior, and the differentiated reward stays closest to the FT prior on $q_\text{A}$.

\begin{table}[h!]
\caption{ESM-2 (150M) pseudo-perplexity (pPPL) on the top-$30$ composition-scoring sequences from each policy. Lower is more natural-protein-like.}
\label{tab:topk}
\vskip 0.05in
\begin{center}\begin{small}
\begin{tabular}{lrrrr}
\toprule
 & \multicolumn{2}{c}{$q_\text{A}$} & \multicolumn{2}{c}{$q_\text{B}$} \\
\cmidrule(lr){2-3}\cmidrule(lr){4-5}
Source & pPPL$\downarrow$ & $\bar L$ & pPPL$\downarrow$ & $\bar L$ \\
\midrule
base ProtGPT2 & 4.52 & 352 & 4.23 & 360 \\
FT only & 6.13 & 349 & 6.39 & 258 \\
RL global-deviation & 15.42 & 462 & 7.24 & 431 \\
RL cosine & 9.80 & 364 & 6.85 & 466 \\
RL symmetric & 17.00 & 440 & 10.76 & 489 \\
RL differentiated & 8.48 & 307 & 12.29 & 403 \\
\bottomrule
\end{tabular}
\end{small}\end{center}
\vskip -0.1in
\end{table}

\section{iid floor: length dependence}\label{app:iid-length}
Table~\ref{tab:iid-length} reports ESM-2 pPPL of the composition-only iid baseline at $L\in\{100,250,400\}$ AA on both targets. The floor shrinks by less than $1$ pPPL unit between $L{=}100$ and $L{=}400$ on each target, supporting the use of the $L{=}400$ value in Table~\ref{tab:quality} as an approximately length-independent floor for the RL policies (whose mean lengths fall inside this range, Table~\ref{tab:esm2}).

\begin{table}[h!]
\caption{ESM-2 (150M) pseudo-perplexity (pPPL) of the composition-only iid floor at three lengths, per target ($n{=}100$ sequences each, residues sampled iid from the corresponding target AA distribution; mean$\pm$std).}
\label{tab:iid-length}
  \begin{center}
    \begin{small}
      \begin{sc}
        \begin{tabular}{lcc}
            \toprule
            $L$ & iid pPPL ($q_\text{A}$) & iid pPPL ($q_\text{B}$) \\
            \midrule
            $100$ & $18.97\pm 1.06$ & $19.27\pm 1.14$ \\
            $250$ & $18.46\pm 0.74$ & $18.77\pm 0.80$ \\
            $400$ & $18.07\pm 0.57$ & $18.50\pm 0.55$ \\
            \bottomrule
        \end{tabular}
      \end{sc}
    \end{small}
  \end{center}
  \vskip -0.1in
\end{table}

\section{Diversity pulse and re-injection}\label{app:divboost}

The diversity-pulse mechanism introduced in \S\ref{sec:method:rl} is parameterized as follows. The baseline candidate pool is sampled at temperature $T{=}0.80$ and $\mathrm{top\text{-}p}{=}0.90$, and the diversity-pulse pool is sampled at $T{=}0.88$ and $\mathrm{top\text{-}p}{=}0.93$ (i.e.\ both knobs slightly hotter). Both pools pass through the same length and MMseqs2 identity filters before merging. Two fractions schedule the merge across training:
\begin{itemize}
\item \textbf{Re-inject fraction} (carry-over of merged pool to the next iteration's prompt set): $0.35$ for the first $20\%$ of iterations, $0.45$ during the early-boost window, and $0.55$ during the late ($\geq 80\%$) window.
\item \textbf{Pulse mix-in fraction} (share of the diversity-pulse pool added to the baseline pool inside one iteration): $0.20$, $0.30$, and $0.35$ on the same three windows.
\end{itemize}
Hotter decoding alone would degrade reward. The re-injection schedule is what turns the diversity pulse into a useful signal, and the schedule is identical across all $160$ runs reported in the main text.

\begin{figure}[h!]
  \vskip 0.2in
  \begin{center}
\centerline{\includegraphics[width=0.6\columnwidth]{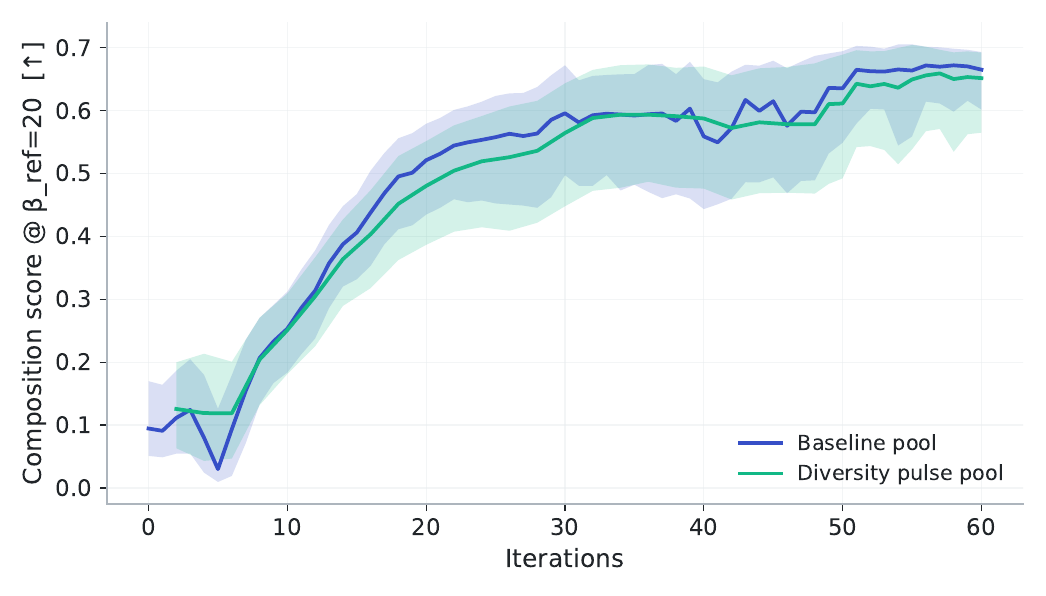}}
\caption{Per-iteration composition score (re-scored at fixed $\beta_\text{ref}=20$ with the \emph{differentiated} variant) of the baseline pool versus the diversity-pulse pool, aggregated across the top $6$ \emph{differentiated}-variant seeds on $q_\text{A}$. Lines: per-iteration median; shading: per-iteration $25$-$75\%$ inter-quartile range across all surviving candidate sequences. The pulse pool tracks the baseline pool throughout training, lagging it slightly during the contraction regime (roughly the first $30$ iterations) and converging back to it afterwards.}
\label{fig:trajectories}
\end{center}
\end{figure}

Figure~\ref{fig:trajectories} summarizes the per-iteration composition score (re-scored at fixed $\beta_\text{ref}=20$) of sequences sampled by the two decoders, aggregated across the top $6$ \emph{differentiated} composition seeds on $q_\text{A}$ (matching the seed selection used in Fig.~\ref{fig:dynamics}). The diversity-pulse pool tracks the baseline pool throughout training, lagging it by no more than a few hundredths of a composition-score unit during the middle of training and converging back to it once the policy has contracted around the target. The pulse, therefore, acts as a controlled exploration channel. It does not collapse reward relative to the baseline pool, but it keeps a wider per-sequence distribution available to the reward-weighted update during the regime where the policy is still moving.

\end{document}